\documentclass[letterpaper, 10 pt, conference]{ieeeconf}  

\IEEEoverridecommandlockouts                              

\usepackage{etextools}
\usepackage{amssymb}
\usepackage{float}
\usepackage{tabto}
\usepackage{makeidx}
\usepackage{amsmath}
\usepackage{bbm}
\usepackage{dsfont}
\usepackage{epsfig}
\usepackage{epsf}
\usepackage{psfrag}
\usepackage{verbatim}
\usepackage{color}
\usepackage{multirow}
\usepackage{tabularx}
\usepackage{booktabs}
\usepackage[tight,footnotesize]{subfigure}
\usepackage{array}
\usepackage{soul}
\usepackage{footnote}
\usepackage{cite}
\usepackage{dblfloatfix}
\usepackage{changepage}

\usepackage{color, colortbl}
\usepackage[colorlinks,bookmarksnumbered,citecolor=orange,urlcolor=orange]{hyperref}
\usepackage{graphicx}
\graphicspath{{Figures}}
\DeclareGraphicsExtensions{.pdf,.png}
\usepackage{bigstrut}
\usepackage[english]{babel}

\usepackage{csquotes}

\usepackage[T1]{fontenc}
\usepackage{algorithm, setspace}
\usepackage{algpseudocode}
\usepackage{url}
\usepackage{multirow}
\usepackage{float}
\usepackage{xcolor}
\usepackage{hyperref}
 \hypersetup{
     colorlinks=true,
     linkcolor=orange,
     filecolor=orange,
     citecolor=orange,      
     urlcolor=orange,
     }

\newcommand{\p}[1]{\smallskip \noindent \textbf{{#1}.}}
\newcommand{\eq}[1]{Equation~(\ref{eq:#1})}

\usepackage{balance}

\title{\LARGE \bf
On the Feasibility of A Mixed-Method Approach for Solving Long Horizon Task-Oriented Dexterous Manipulation
}

\author{Shaunak A. Mehta$^{1,*}$ and Rana Soltani Zarrin$^{2}$
\thanks{*This work was completed during an internship at Honda Research Institute USA.}
\thanks{$^{1}$ Department of Mechanical Engineering, Virginia Tech, Blacksburg, VA 24060.
        {\tt\small mehtashaunak@vt.edu}}%
\thanks{$^{2}$ Honda Research Institute,
        San Jose, CA 95134.
        {\tt\small rana\_soltanizarrin@honda-ri.com}}%
}

\begin{document}

\maketitle
\thispagestyle{empty}
\pagestyle{empty}

\begin{abstract}
In-hand manipulation of tools using dexterous hands in real-world is an underexplored problem in the literature. 
In addition to more complex geometry and larger size of the tools compared to more commonly used objects like cubes or cylinders, task oriented in-hand tool manipulation involves many sub-tasks to be performed sequentially. This may involve reaching to the tool, picking it up, reorienting it in hand with or without regrasping to reach to a desired final grasp appropriate for the tool usage, and carrying the tool to the desired pose. 
Research on long-horizon manipulation using dexterous hands is rather limited and the existing work focus on learning the individual sub-tasks using a method like reinforcement learnNing (RL) and combine the policies for different subtasks to perform a long horizon task. However, in general a single method may not be the best for all the sub-tasks, and this can be more pronounced when dealing with multi-fingered hands manipulating objects with complex geometry like tools. In this paper, we investigate the use of a mixed-method approach to solve for the long-horizon task of tool usage and we use imitation learning, reinforcement learning and model based control. We also discuss a new RL-based teacher-student framework that combines real world data into offline training. We show that our proposed approach for each subtask outperforms the commonly adopted reinforcement learning approach across different subtasks and in performing the long horizon task in simulation. Finally we show the successful transferability to real world.


\end{abstract}


\section{Introduction}

Imagine trying to pick up a wrench and positioning it in your hand to tighten a bolt. Owing to the high dimensional degrees of freedom and contact configurations, multi-fingered robot hands seem capable of performing a variety of such in-hand manipulation tasks. However, enabling dexterous robot hands to perform those tasks in the real world is an ongoing research. Some of the associated challenges include determining appropriate grasps for different phases of the task such as pickup to reorientation to tool-use and robustly realizing the (re)grasps on these complex object shapes, and successful realization of the plans in the real world considering uncertainties, errors, and Sim2Real gaps. 

Due to the challenges in planning and control, existing research \cite{kadalagere2023review} has explored deep learning approaches for dexterous grasping \cite{zhao2020grasp} and in hand object reorientation \cite{chen2023visual}, mainly for objects with simple geometry \cite{openAI2020learning, Matei2022gaiting}. 
There has been limited exploration on solving for long-horizon tasks using dexterous hands which involve object pick up and regrasping while simultaneously reorienting and moving the object to a desired location in space for further usage. Long-horizon manipulation tasks are especially challenging when learned with a single approach for all subtasks \cite{chen2023sequential}, and may lead to a low success rate in deployment. Our insight is that

\begin{center}
    \textit{
    Different segments of a long-horizon task can be solved using different approaches for increased success rate
    }
\end{center}

\begin{figure}
    \centering
    \includegraphics[width=\columnwidth]{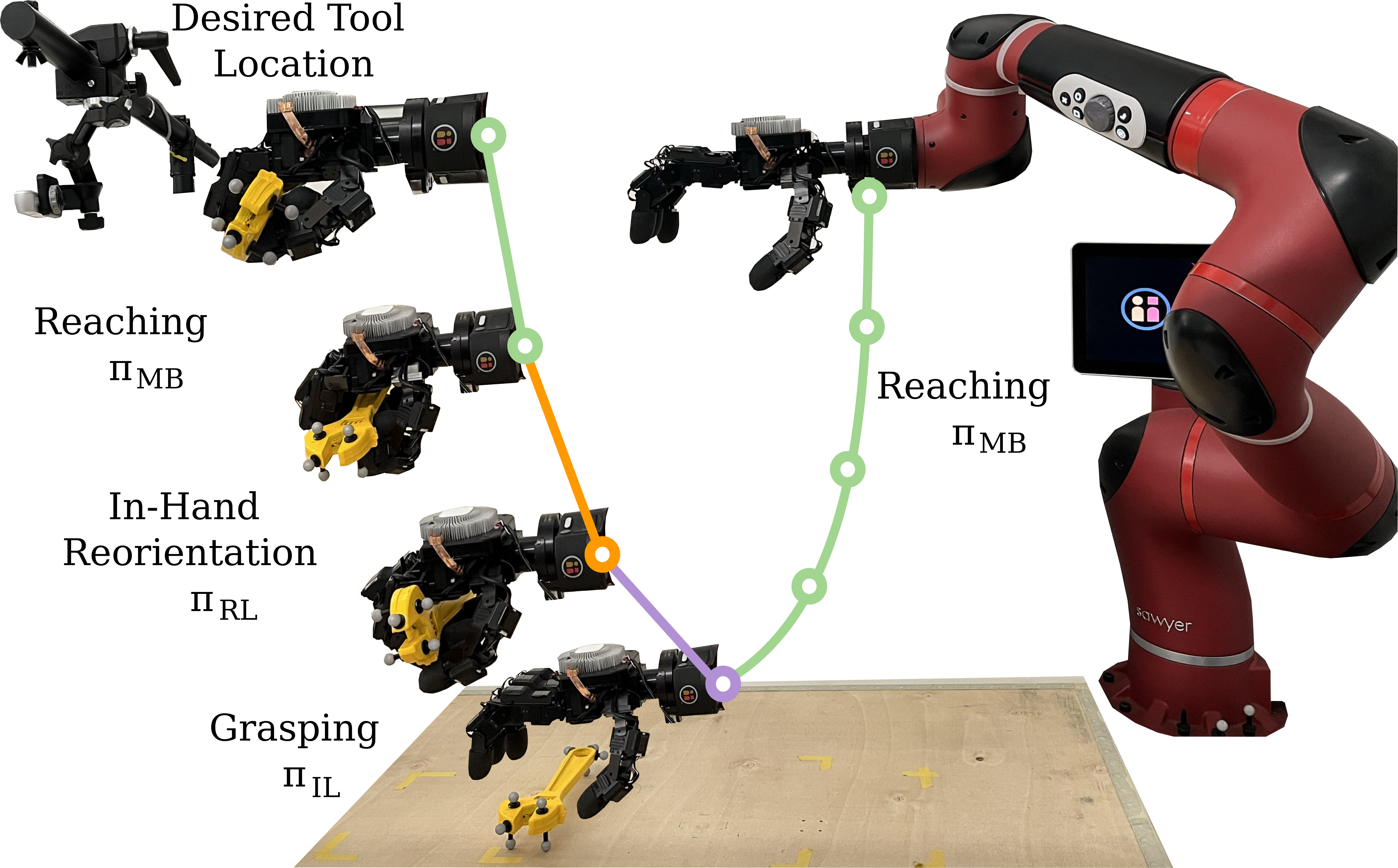}
    \caption{The execution of a long-horizon dexterous manipulation task. The robot arm first reaches for the tool using model-based policy $\pi_{MB}$, followed by grasping using a policy trained using imitation learning $\pi_{IL}$. The robot then performs in-hand manipulation using a reinforcement learning policy $\pi_{RL}$ and then using the $\pi_{MB}$ to carry the tool to the desired location.}
    \vspace{-1.5em}
    \label{fig:front}
\end{figure}

Using this insight, we propose combining different policies under a single framework, where each segment of a long-horizon task is performed using a method better suited for the subtask.
Such a method is determined based on the conditions of each task in terms of available information and the level of human effort required in terms of defining the models, providing demonstrations, and designing the reward functions for different approaches of solving for each subtask.
We then combine the different methods to perform a long-horizon task using a higher level planner that determines the appropriate low-level policy to be executed.




Our contributions are as follows: we propose (1) using a mixed method approach for solving long-horizon tasks where each subtask is solved using a method better suited to that subtask. The low-level policies are then combined under a unified framework and selected for rollout through a higher-level policy; (2) a novel teacher-student reinforcement learning approach which incorporates sparse real-world data into offline training in simulation. 

We perform ablation studies for each individual subtasks comparing the performance of our proposed method to standard reinforcement learning. We then compare the overall performance of long-horizon task completion for our unified approach and sequential execution of low-level policies trained using reinforcement learning in a simulated environment. Finally, we show that our approach can achieve a high success rate for a long-horizon task involving tools when deployed in the real world. 



\section{Related Works} \label{sec:related}
In this section, we summarise the recent works in the domain of dexterous manipulation as pertained to the long-horizon manipulation.

\subsection{Grasping}
Most of the works in in-hand dexterous manipulation assume that the robot starts with the object in hand. However, in practical applications the robot needs to actually grasp and pick up the object before performing in-hand manipulation. Several works focus on determining the feasible grasp locations and grasp sequences that can be used to successfully pickup the object \cite{wu2024unidexfpm, turpin2022grasp}. These works require an accurate model of the object \cite{wang2023dexgraspnet} and methods for sampling \cite{newbury2023deep} as well as metrics for selecting contact points on the continuous mesh of the object \cite{kopicki2014learning}. While Sim2Real gaps and localization errors makes zero-shot transfer of the simulation- or model-based grasps to realworld difficult, the challenges are even worse for online contact point (re)generation due to computational cost of search-based methods. Learning from human demonstrations has been an alternative approach to overcome the above mentioned challenges \cite{wu2023learning, huang2023dexterous}. Other works use human demonstrations for guiding the search-based or learning-based methods \cite{zarrin2023hybrid, gordon2023online} in combination with grasp quality metrics to facilitate the process. Nevertheless, real-world deployment suffer from challenges such as object occlusion, or difficulty of precisely determining the contacts points on the hand and the object when the hand makes contact with the object. In our framework, we propose using an imitation learning based approach, that 
is not prone to occlusions and does not require object models, for learning to grasp an object from human provided demonstrations.  


\subsection{In-Hand Reorientation} 
In the past, research has focused on a wide spectrum of approaches --- from model-based control to imitation and reinforcement learning ---  to solve for the complex task of in-hand manipulation of objects \cite{kadalagere2023review}. 
\cite{cheng2023enhancing, dafle2014extrinsic} try to define precise models and contact dynamics to solve the task of in-hand object reorientation for simple geometric objects. While these works achieve success in transferring the model based approach to real world, it may not be possible to design the contact dynamics and physical parameters of irregularly shaped tools into a model. 
To tackle the complexity in accurately modelling the dynamics of the real world, reinforcement and imitation learning are used as alternatives.
On one hand, imitation learning has been studied to perform dexterous manipulation tasks \cite{qin2022dexmv, arunachalam2023holo}. Teleoperation approaches developed to control the anthropomorphic hands are leveraged to collect the demonstrations \cite{handa2023dextreme, qin2023anyteleop}. Even if teleoperation can be used to perform simple tasks, it may not be intuitive for a user to provide demonstrations for complex reorientation tasks, without force feedback during teleoperation.
On the other hand, following OpenAI's \cite{openAI2020learning} work on reinforcement learning for in-hand dexterous manipulation, subsequent research works have focussed on learning dexterous manipulation policies from visual feedback \cite{ handa2023dextreme, allshire2022transferring}, tactile feedback \cite{sievers2022learning} and a combination of tactile and visual feedback \cite{qi2023general, yuan2023robot}.
While reinforcement learning paired with domain randomization \cite{tobin2017domain} may make the policy robust to noise, generated behaviors may not always work in real world. This issue is especially prominent in the manipulation of irregularly shaped objects like tools, resulting due to the Sim2Real physics gaps. In this paper, we propose an approach for teacher-student based reinforcement learning in which the student has access only to a subset of information accessible by the teacher, while having access to real-world data incorporated into the offline training.

\subsection{Long-Horizon Tasks}
Even though the essence of performing long-horizon tasks is to learn a policy that encapsulates the different tasks such as grasping and in-hand manipulation defined above, the research in this domain for dexterous manipulation tasks is rather limited. The problem of performing long-horizon tasks is often solved by breaking down the long-horizon task into multiple subtasks and learning a policy for each of these subtasks individually using hierarchical learning or sequential learning \cite{barto2003recent, gupta2021reset, mehta2024waypoint}. Similar approaches are explored for dexterous manipulation, where reinforcement learning is used to train low level policies for the different subtasks and a higher level policy is used to determine the policy to be used \cite{li2024interactive} or to determine the policy switching probabilities to enable smooth transitioning between the different policies \cite{chen2023sequential}. While \cite{li2024interactive, chen2023sequential} show the feasibility of their approach for tool manipulation, the experimental validation of their approach for tool manipulation is limited to simulated environments. Our work is most similar to \cite{chen2023sequential}, in the sense that we break down the task into multiple subtasks and use a high level policy to determine the lower level policy to be used. However, instead of relying solely on reinforcement learning to learn the low level policies, we utilize the best suited approach to solve for each subtask --- imitation learning, reinforcement learning or model based control. We show that our proposed approach achieves higher rewards and success rate for individual sub-tasks compared to RL, and that the policy can be successfully transferred to the hardware with a high success rate. 
\section{Problem Statement} \label{sec:problem}
\begin{figure}
    \centering
    \includegraphics[width=1\columnwidth]{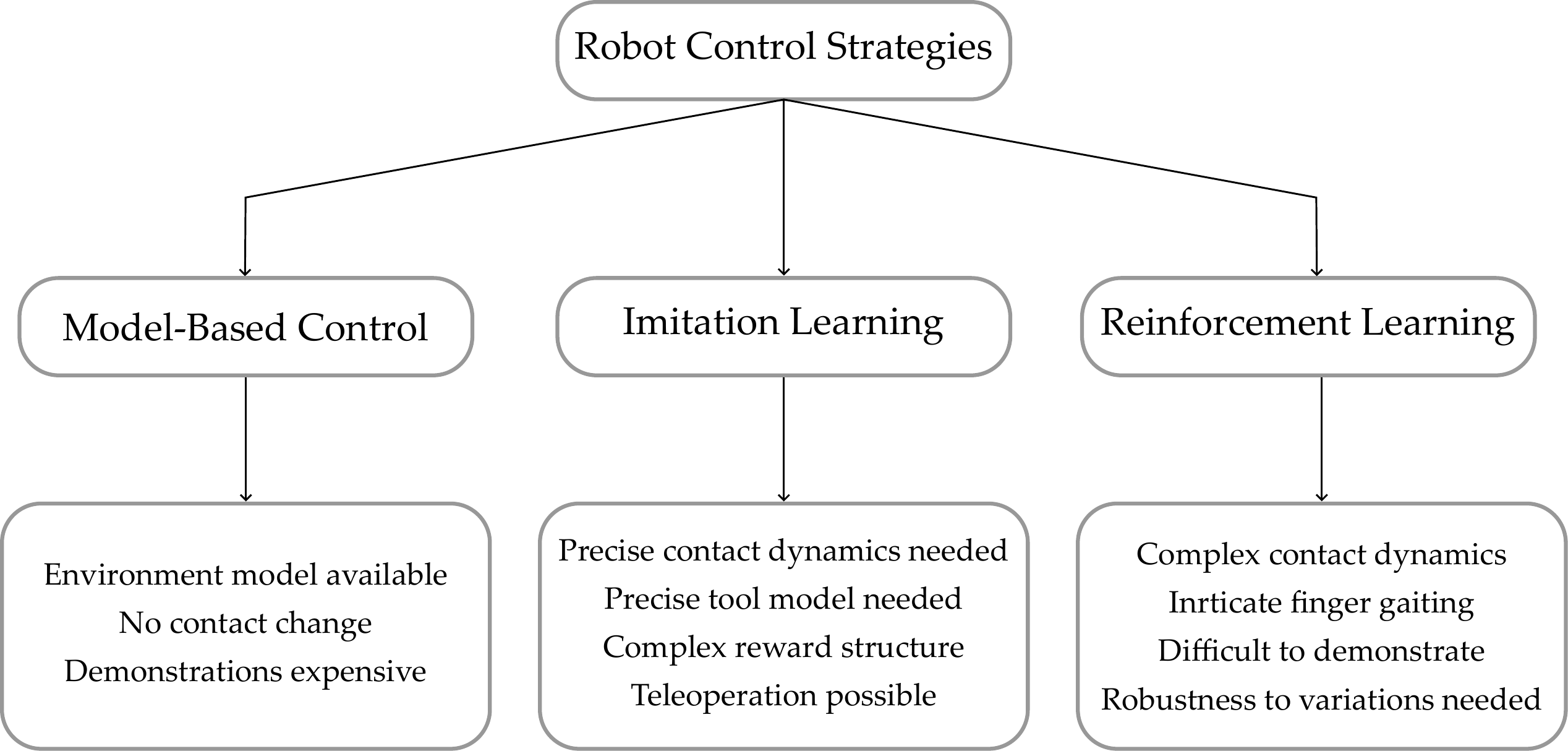}
    \caption{Choosing a suitable method for solving a subtask. The flowchart highlights the different robot control strategies that can be used to solve a given subtask in a long-horizon dexterous manipulation task. It further provides the conditions for employing each method.}
    \vspace{-1.5em}
    \label{fig:designer}
\end{figure}

We consider scenarios where the robot hand has to reach for a tool, grasp it and perform in-hand reorientation to hold the tool in a feasible position for use. Such a task is illustrated in Figure \ref{fig:front}. We break down this task into three subtasks: 
\begin{itemize}
    \item \textbf{Reaching} for or carrying the tool to the desired location
    \item \textbf{Grasping} and picking up the tool while regrasping if necessary to move to a more stable grasp
    \item \textbf{Reorienting} the tool in hand to a desired orientation
\end{itemize}
Traditional way of solving for this problem has been using a single approach to solve for the different sub-tasks. However, as discussed in section \ref{sec:related}, not all the sub-tasks can be best solved by the same approach. 
In this section, we propose the conditions for different strategies to better solve for each of the subtasks, and summarize these conditions in Figure \ref{fig:designer}.



In our running example from Figure \ref{fig:front}, for a simple subtask like reaching for a tool and carrying the tool to a desired location, planning and control methods can be developed without the consideration of complex finger gaiting or contact dynamics. In such tasks, where accuracy is of importance and a model is readily available, using reinforcement learning to learn a policy may lead to a policy that is suboptimal or may have noise in reaching the desired pose. Also, a new policy needs to be trained for any change in the environment, making the approach computationally expensive. On the other hand collecting human demonstration data for such a trivial task may be expensive and time consuming. To that end, for task segments that do not involve hard to model contact dynamics, model based planning and control approaches \cite{gasparetto2015path} can be easily employed that can accommodate changing environments without the need for learning. In this paper, we use model based trajectory optimization to solve for the \textbf{Reaching} tasks.

For the \textbf{Grasping} subtask, a model-based control approach requires defining a precise dynamics model for the tool and the robot hand. It also needs the precise information about the contact dynamics which may be difficult to model. On the other hand, reinforcement learning based approaches need careful design of the reward function to enable the robot to learn to grasp the tool in a stable and legible manner with smooth actions that can enable feasible real-world implementation. The design of this dense reward function requires an expert fine tuning the reward function over multiple iterations to achieve the desired grasping behavior. Instead, if an expert can provide a few  demonstrations, showing the robot hand how to grasp and pick up the tool, the robot can learn the task successfully using imitation learning. Unlike RL which may generate actions not successfully transferable to hardware, the actions  provided by an expert user and retargetted to the robot hand will lead to a successful grasp while being smooth and legible. In our framework, we propose using imitation learning for grasping tasks, that can lead to successful learning in a shorter time duration with limited computational resources. 

Finally for the in-hand \textbf{Reorientation} subtasks, similar to the \textbf{Grasping} subtask, using a model-based control approach requires precise dynamic modelling and contact information, making the model-based control approach infeasible for the complex reorientation task. 
Also, to enable the robot to reorient the tool to any desired position using imitation learning, it may be difficult for the operator to provide good demonstrations due to object occlusion and lack of accurate visual feedback. Due to this, 
a large number of demonstrations may be needed,
making the approach expensive in terms of human time and effort. On the other hand, similar to the approaches described in the Section \ref{sec:related}, when the tool is already stably in the robot's hand, the robot can learn to reorient the tool to the desired position with reinforcement learning without the need for complex fine tuning of the reward functions. In this paper, we propose a novel teacher-student approach for reinforcement learning for solving the in-hand reorientation task.

We define the problem of solving a \textbf{Long Horizon Task} as a Markov Decision Process (MDP) $\mathcal{M} = \langle \mathcal{S}, \mathcal{A}, T, r, H \rangle$, where $s^t \in \mathcal{S}$ is the state of the world and $a^t \in \mathcal{A}$ is the action taken by the robot at timestep $t$. The robot transitions to the next state $s^{t+1}$ according to the transition function $T(s^t, a^t)$.
At each timestep, the robot receives a reward from the environment defined by the reward function $r:s^t, a^t \rightarrow \mathbb{R}$ and the interaction ends after a maximum of $H$ timesteps.
\section{Method} \label{sec:method}
Our insight is that using different algorithms for solving for different segments of a long horizon task can lead to high success rate of task execution in real world deployment.
In Section \ref{subsec:imitation}, we outline an approach of imitation learning to solve for \textbf{Grasping} and pickup tasks. Next, in Section \ref{subsec:rl}, we propose a novel teacher-student framework for reinforcement learning that learns in-hand \textbf{Reorientation} by incorporating the real world data to enable a smooth sim2real transfer. Section \ref{subsec:model}, discusses the model-based control approach for solving for the \textbf{Reaching} tasks. Finally, in Section \ref{sec:high-level}, we introduce the framework that combines all these approaches to solve for long horizon dexterous manipulation tasks.

\subsection{Imitation Learning: Ensemble Networks with Look Ahead}\label{subsec:imitation} 
\begin{figure}
    \centering
    \includegraphics[width=1\columnwidth]{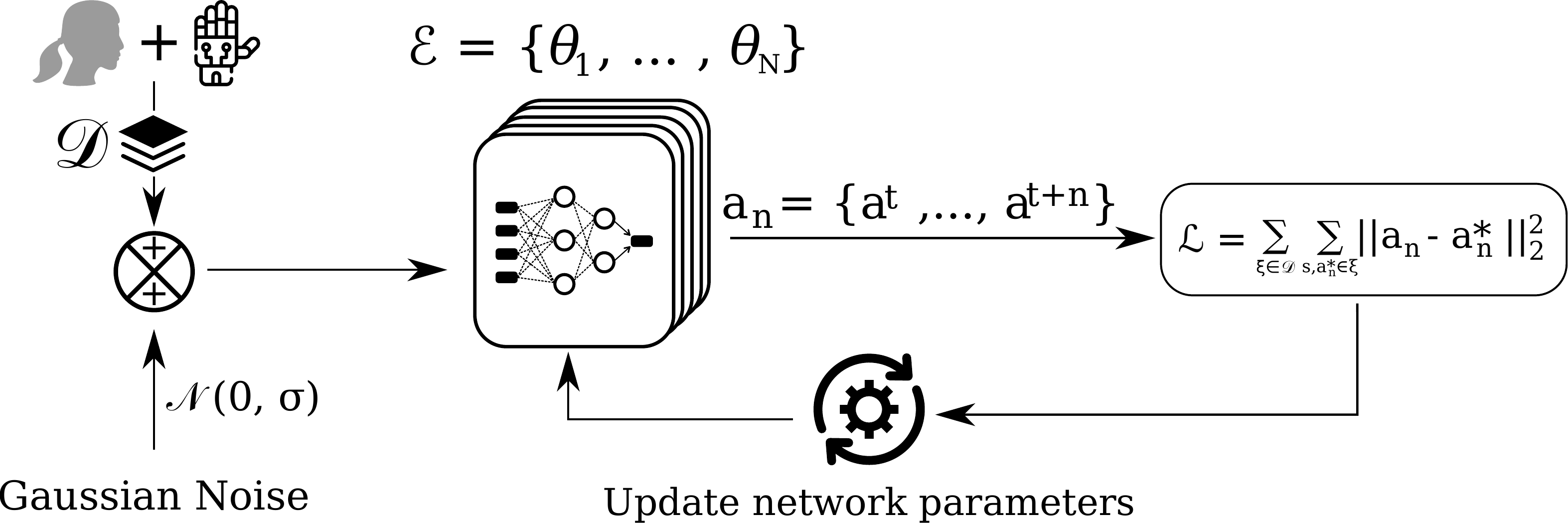}
    \caption{Imitation Learning Framework. In this approach, the gaussian noise is added to the human demonstrations to augment the data and an ensemble of $N$ networks is trained. Each network is trained independently to predict a series of $n$ actions in the future from a given state $s$.}
    \vspace{-1.5em}
    \label{fig:imitation}
\end{figure}

Our imitation learning framework for grasping and pick up tasks using dexterous hands is depicted in Figure \ref{fig:imitation}.
In imitation learning, we assume access to a set of expert provided demonstrations $\mathcal{D} = \{\xi_1, \xi_2, \cdots\}$, where $\xi_i = \{s_i^0, a_i^0, \cdots s_i^H, a_i^H\}$. These demonstrations can be provided using any available off-the-shelf teleoperation approach for dexterous manipulatiopn \cite{handa2023dextreme, qin2023anyteleop}, but for our implementation we use the teleoperation framework defined in \cite{arunachalam2023holo}. 
We then add zero mean gaussian noise ($\mathcal{N}(0, \sigma)$) to the demonstrations to make the dataset more diverse and the imitation learning policy robust to noise in the system. Following the findings of \cite{belkhale2024data}, we only add noise to states of the demonstrations while not altering the actions or the target states. 

We use this dataset of augmented demonstrations to train a policy to grasp and pick up an object using imitation learning. In order to reduce the noise and the distribution shift during deployment, we train an ensemble of $N$ policies with weights $\mathcal{E} = \{\theta_1, \theta_2 \cdots \theta_N\}$ \cite{menda2019ensembledagger}. Each policy is trained to predict $n$ actions in the future (look-ahead) \cite{chi2023diffusion} by minimizing the Mean-Squared-Error loss function defined as
\begin{equation}\label{eq:m0}
    \mathcal{L} = \sum_{\xi \in \mathcal{D}}\sum_{s,a_n^* \in \mathcal{\xi}} \|a_n - a_n^*\|^2
\end{equation}
where $a_n = \{a^t, a^{t+1} \cdots a^{t+n}\}$ is the set of predicted actions, $a_n^*$ is the set of optimal actions from a given state $s$ and $\|(\cdot)\|$ represents the $L2$ norm.

Each of the policies in the ensemble is a fully connected multilayer perceptron with $5$ hidden layers and rectified linear activation units. Our ensemble includes $10$ independently trained policies, where each policy is initialized with a random seed and and trained to optimize the network weights using the Adam optimizer with a learning rate of 0.001. 
On deployment, we use the ensemble of policies $\mathcal{E}$ to predict a set of $N$ actions and the average of these $N$ actions is used as a control input to the robot to minimize the uncertainty in action prediction. We note that since this framework only depends on the state of the system and not on the visual feedback, the learned policy is not affected by visual occlusions of the hand or the object.

\subsection{Reinforcement Learning: Teacher-Student Framework Incorporating Real World Data}\label{subsec:rl}
Teacher-student frameworks have been used in the past to learn in-hand object reorientation \cite{chen2023visual, qi2023general, yuan2023robot}, where the teacher is trained with privileged information (information not available in real world), and the student's observation space is made sparse while using domain randomization \cite{tobin2017domain} and the teacher's actions to learn a robust reinforcement learning policy. In our proposed framework, instead of just relying on domain randomization and the teacher policy for training the student, we collect some data in the real world and incorporate it in the student's learning framework to make the policy robust for real world tuning and deployment. 

\subsubsection{Teacher Policy} 
\begin{figure*}
    \centering
    \includegraphics[width=1.8\columnwidth]{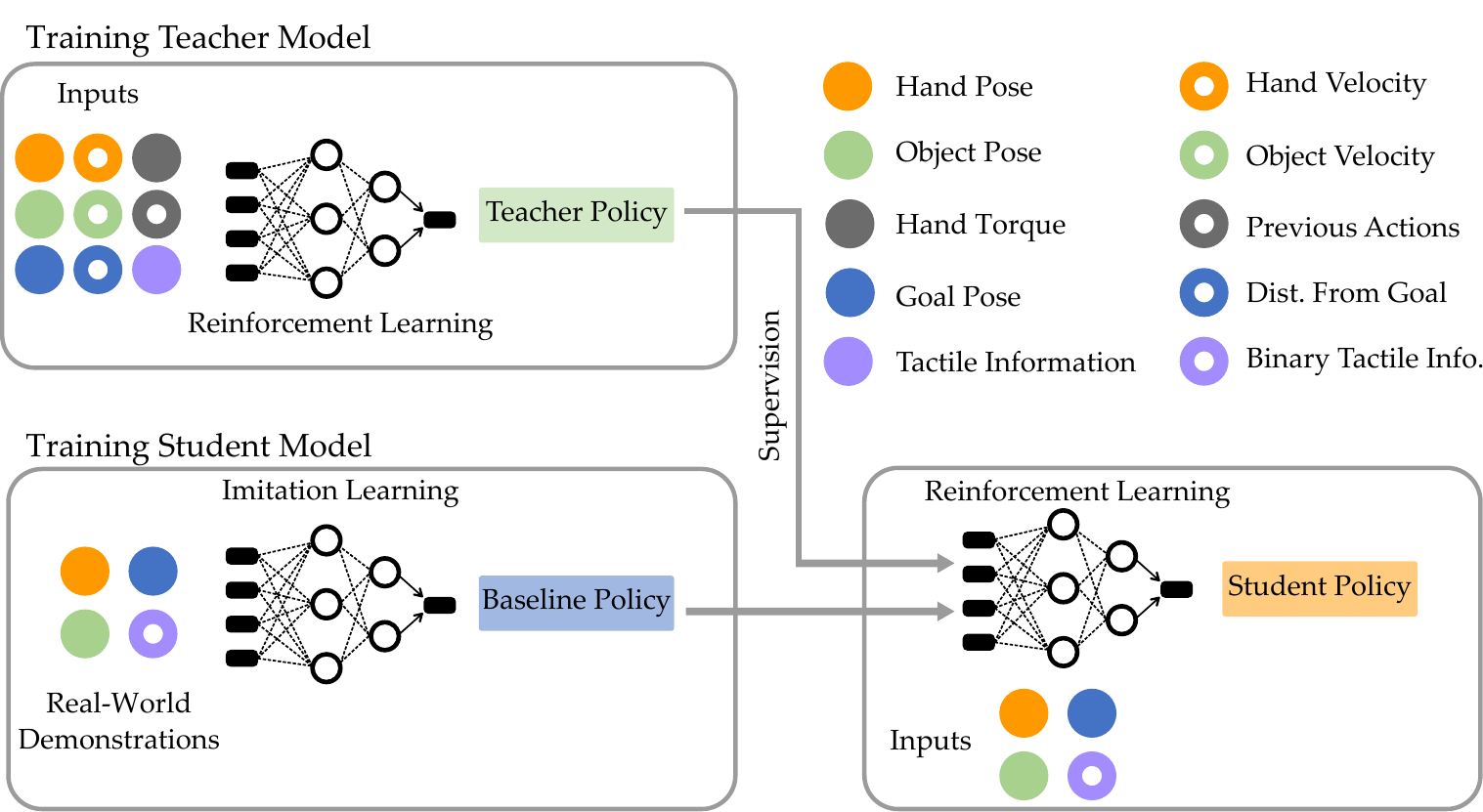}
    \caption{Teacher-Student Framework Incorporating Real World Demonstrations. The teacher model is trained with privileged information that may not be easily available in the real world.. This teacher model is then used to supervise the training of the student model which is pretrained from demonstrations provided by an expert in the real world. Finally, the student model is deployed in the real world for fine-tuning in the real world without supervision to adapt to the hardware control and physical parameters.}
    \vspace{-1.6em}
    \label{fig:student-teacher}
\end{figure*}

The learning of teacher policy $\pi^T$ is framed as a reinforcement learning problem where the teacher observes the state of the world $s^t$ at a timestep $t$, takes an action $a^t$ and receives a reward $r(s^t, a^t)$ from the environment. The policy is trained using proximal policy optimization (PPO) \cite{schulman2017proximal} to maximize the expected discounted return of the episode $\pi_{T^*} = \arg\max_{\pi_T} \mathbb{E} \sum_{t=1}^H \gamma^t \cdot r(s^t, a^t)$, where $\gamma$ is the discount factor. The teacher model's observation space includes privileged information that is not easily available in real world, but accessible in the simulation. This privileged information includes precise tool and hand joint position and velocity, hand joint torques, tactile force information and feature information for the task as shown in Figure \ref{fig:student-teacher}. The reward function for training this privileged teacher model is given as 
\begin{align}
    r_T(s^t, a^t) &= \alpha_1 \dfrac{1}{\|\Delta \theta^t\| + \epsilon_\theta} \text{dense task reward}\\
    &+ \alpha_2 |\dot{q}^t - \dot{q}^{t-1}| \text{action penalty}\\
    &+ \alpha_3 \|\tau^t\| \text{torque penalty}\\
    &+ \alpha_4 \mathds{1} \text{   (Task success)} \text{Success bonus}\\
    &+ \alpha_5 \mathds{1} \text{   (Tool dropped)} \text{Failure penalty}
\end{align}
where $\Delta \theta^t$ is the distance from the desired tool orientation, $q^t$ represents the joint states of the robot hand, $\tau^t$ is the torque applied by the joints and $\mathds{1}$ is an indicator function. $\alpha_1, \alpha_4, \epsilon_\theta > 0$ and $\alpha_2, \alpha_3, \alpha_5 < 0$ are constants that determine the relative weight of terms in the reward function. For our implementation, we used $\alpha_1 = 1.0, \alpha_2 = -0.1, \alpha_3 = -0.01, \alpha_4 = 250, \alpha_5 = -100$ and $\epsilon_\theta = 0.001$.

\subsubsection{Student Policy}

Now that we have outlined the observation space and the reward function used to train the teacher model, in this section, we discuss in detail the training of the student model. As we discussed earlier, the student model is trained with data that we can easily obtain in the real world. To that end, the observation space of the student model is a subset of the observation space of the teacher which includes the hand and tool pose, the goal pose and binary tactile information as shown in Figure \ref{fig:student-teacher}.\\
\p{Pre-training} Now, to incorporate real world data in the training of the student, we start with pre-training the student model with $25$ demonstrations for the reorientation task collected in the real world. Similar to Section \ref{subsec:imitation} demonstrations are collected using the teleoperation framework defined in \cite{arunachalam2023holo}. The imitation learning policy is then trained using the framework described in Section \ref{subsec:imitation}, with no look ahead data (i.e. $n = 1$). We note that even though these demonstrations may not be enough to learn accurately the complex task of reorientation, it initializes the student model with a baseline policy including real world data.\\
\p{RL with teacher supervision} Next, we build on the baseline policy trained using imitation learning and real world data to train the student model using reinforcement learning. Similar to the teacher model, we train the student model using PPO to maximize the expected return of the episode $\pi_{S-T} = \arg\max_{\pi_{S-T}} \mathbb{E} \sum_{t=1}^H \gamma^t \cdot r(s^t, a^t)$. The student with a sparse observation space and the teacher supervision is trained using the following reward function 
\begin{align}
    r_S(s^t, a^t) &= \alpha_1 \dfrac{1}{\|\Delta \theta^t\| + \epsilon_\theta} \text{dense task reward}\\
    &+ \alpha_2 \|a^{t}_T - a^{t}_S\| \text{teacher supervision} \\
    &+ \alpha_3 \mathds{1} \text{   (Task success)} \text{Success bonus}\\
    &+ \alpha_4 \mathds{1} \text{   (Tool dropped)} \text{Failure penalty}
\end{align}
where $a_T$ and $a_S$ are the teacher and student actions respectively.  In our implementation, we set $\alpha_1 = 1.0, \alpha_2 = -0.1, \alpha_3 = 250, \alpha_4 = -100$ as the relative weights of the terms in the reward function and $\epsilon_\theta = 0.001$.

\subsection{Model-Based Control} \label{subsec:model}
For this approach, we assume access to a model of the environment and the related constraint parameters in the form of a reward function $r_\theta(s^t)$. This reward function may be hand-crafted or be learned from demonstrations from a human  user \cite{argall2009survey, mehta2023unified}. As we discussed in Section \ref{sec:problem}, for tasks where the environment constraints may change frequently, and where a dynamic model of the environment is readily available, model based approaches can be utilized to generate safe and reliable robot trajectories. In this setting, we leverage the underlying robot kinematics and constrained optimization to solve for the optimal robot trajectory
\begin{equation} \label{eq:m1}
    \xi_r = \arg\max_{\xi \in \Xi}\sum_{s\in \xi} r_\theta(s) \hspace{0.5cm} s.t. \xi(0) = s^0, \xi(H) = s^H
\end{equation}
where $\xi_r$ is the generated robot trajectory, $\Xi$ is the set of all possible trajectories in the environment and $s_0$ and $s_H$ are the start and the goal positions for the robot respectively. There is a wide range of research on trajectory optimization and model-based manipulation for robots that focuses on solving \eq{m1} \cite{gill2005snopt, schulman2014motion}. Our method does not rely on any specific approach or optimizer. For our experiments, we use a sequential least squares programming to solve \eq{m1} to generate the optimal robot trajectory for the task.

\begin{figure}
    \centering
    \includegraphics[width=0.8\columnwidth]{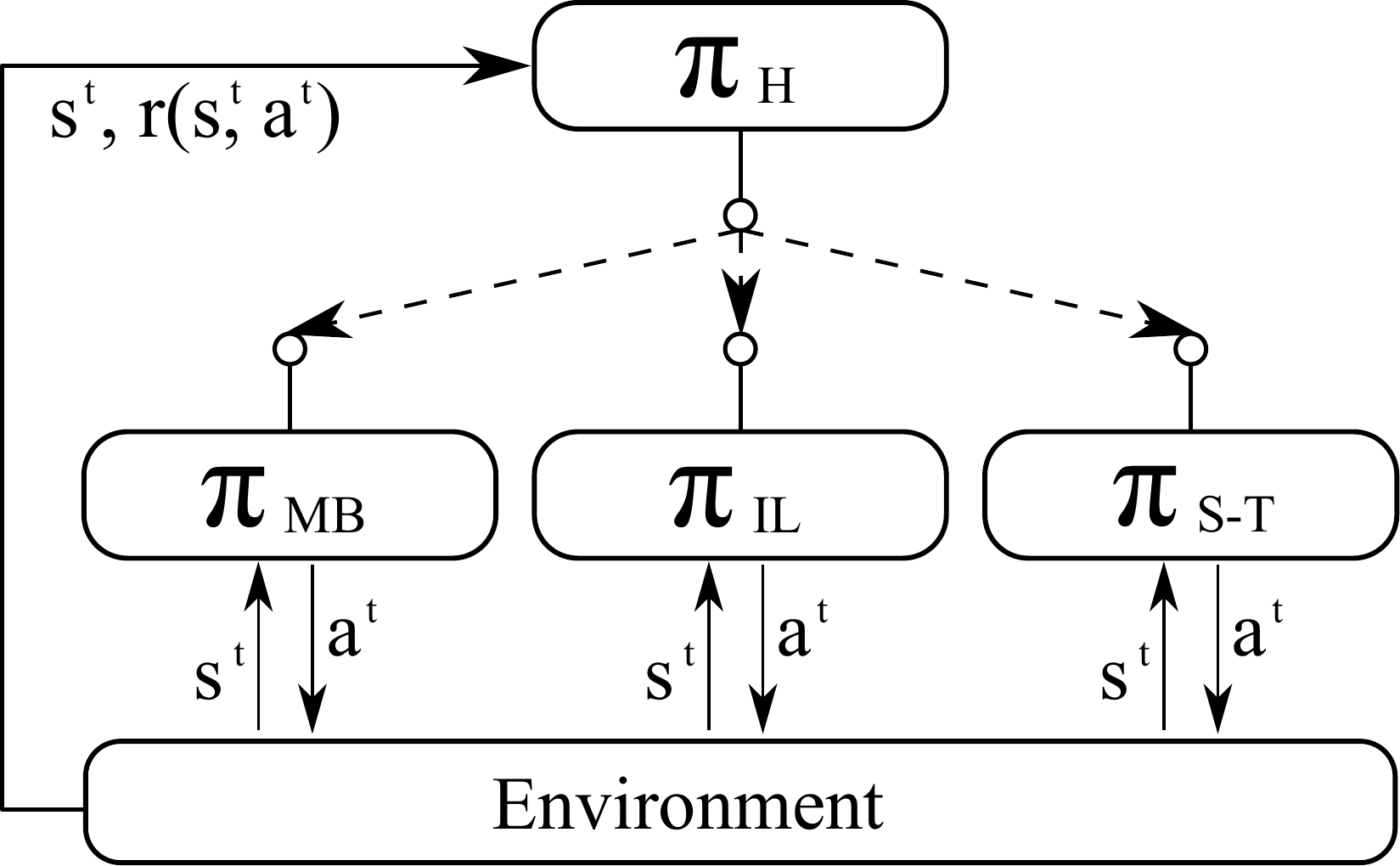}
    \caption{Our proposed unified framework for combining different approaches to solve for a long horizon task. Here the high level policy determines the policy to be used based on the current state of the environment and the lower level policies are used to take action in the environment.}
    \vspace{-2.0em}
    \label{fig:high-level}
\end{figure}

\subsection{A Unified Policy} \label{sec:high-level}
Now that we have defined the different phases of the long-horizon task and the methodologies to be used to solve for each individual phase, in this section, we combine the different policies under a unified framework. We use  a high level policy $\pi_H$ that determines which of the frameworks is to be used based on the current state of the environment. As shown in Figure \ref{fig:high-level}, at each timestep $t$, $\pi_H$ observes the state of the environment (the combination of the observation space of all the other policies)
and outputs the lower level policy to be implemented. The low level policy is then executed and $\pi_H$ receives a reward $r(s^t, a^t)$ from the environment. The objective of $\pi_H$ is to maximize the expected return for the episode. This policy can be trained to maximize this expected return using any standard reinforcement learning algorithm. However, for our implementation, $\pi_H$ is a logical interpreter that determines the framework to be used based on the state of the environment. 
For example, if the robot is within $0.02m$ of the object, the interpreter decides to rollout the imitation learning policy for grasping and picking up the object. Similarly, when the tool is picked up above a height of $0.08m$, the interpreter switches to the in-hand reorientation policy and after reaching within $0.1rad$ of the desired orientation the interpreter switches back to reaching policy to carry the tool to the desired position.
We use this hand-crafted logical interpreter to reduce the overall training time for a single long horizon task. However, it may be useful to use reinforcement learning to train this policy when the number of objects and the tasks increase.

\section{Experiments}

In this section, we perform experiments and evaluate the performance of our proposed framework for solving the long horizon task using a unified approach. First, we compare the performance of each selected sub-task solver with a standard reinforcement learning approach (PPO) and show that our proposed approach in each subtask outperforms the baseline. We then compare the performance of our proposed unified approach that uses the low level policies trained using the approach best suited to each subtask to that of the low level policies trained using only reinforcement learning. Finally, we deploy our proposed unified framework in the real world to test the efficacy of our approach and show that it can successfully complete the long horizon task.
We perform all of our training and simulated experiments in the Isaac Gym \cite{makoviychuk2021isaac} simulator using an Allegro Hand with 22 DoF (4 for each finger + 6 for the base).
At each rollout, we sample the weight and friction parameters of the tool from Gaussian distributions $\mathcal N(100, 20)$ and $\mathcal N(0.7, 0.2)$ respectively. For our real world deployment, we use an Allegro hand mounted on a 7 DoF Sawyer robot. For collecting the binary tactile feedback, we use Xela uSkin\footnote{\href{https://www.xelarobotics.com/}{xelarobotics.com}} patch and fingertip sensors mounted on the Allegro Hand.

\subsection{Reaching Subtask}
In this task, the robot hand needs to reach for and align itself with a tool (wrench) placed in the environment. The position and orientation of the wrench and the hand are randomized in a $0.5 \times 0.5 m$ square at the start of each episode. Each episode lasts for $10$ timesteps and at each timestep, the policy received a reward based on the distance from the goal and sudden changes in velocity (smoothness of actions).  
We compare the performance of the model-based trajectory optimization approach (\textbf{MB}) to that of a standard reinforcement learning (\textbf{RL}) approach and report the reward and success rate for each approach averaged over $25$ rollouts. The task is considered to be a success if the hand is within $0.02m$ and $0.1 rad$ of the object position. Our results show that the model-based trajectory optimization approach outperforms the standard reinforcement learning based policy in terms of the rewards achieved (see Figure \ref{fig:results}). We also observe that given the success criterion of $0.1 rad$, the model based approach had a success rate of $100\%$, while that for RL policy was $36\%$. This result supports our initial hypothesis that a model-based approach is more suitable to solving a task like reaching where a model of the environment is readily available, as compared to reinforcement learning.

\subsection{Grasping Subtask}
Here, the multi-fingered robot hand needs to grasp the tool
and perform finger gating in order to move the tool into a stable grasp (in which the tool can be used to apply force in the environment while maintaining its position in the grasp). For this task, the tool is randomly initialized with gaussian noise $\mathcal{N}(0, 0.05m)$ added to its position, while the start position of the hand is constant. We compare the performance of the policy trained using imitation learning framework discussed in Section \ref{subsec:imitation} (\textbf{IL}) to that of a reinforcement learning policy trained using PPO (\textbf{RL}). 
Each episode lasts for $50$ timesteps and the environment resets at the beginning of each episode. After each timestep, the policy receives a reward based on the finger distance to the tool, the height to which the object is picked up in the grasp, the orientation of the tool and the smoothness of actions. We compute the reward and the success rate of each approach averaged over $25$ rollouts. A grasp is considered successful if the hand lifts the object above a threshold of $0.05m$. We observe that both policies were able to successfully pick up the objects for all rollouts, but the imitation learning policy outperformed the reinforcement learning policy in terms of the reward received (see Figure \ref{fig:results}). Even though the RL policy was able to pick up the tool, it failed to keep the tool upright and took redundant actions leading to a lower reward.  On the other hand the actions taken by the imitation learning policy are more stable and lead the hand grasping the tool in an upright orientation.

\begin{figure}
    \centering
    \includegraphics[width=1\columnwidth]{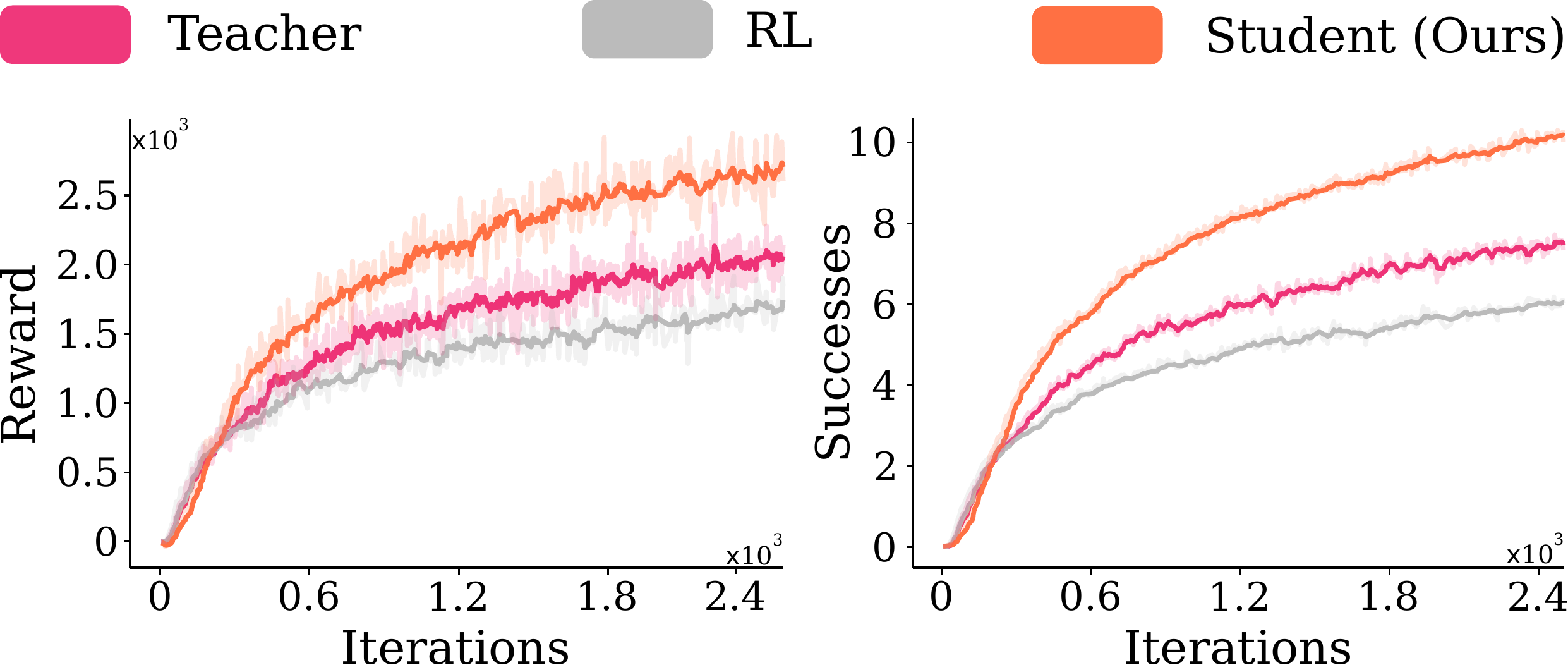}
    \caption{Training curves for our proposed Teacher-Student framework. We observe that the student policy trained using our framework outperforms the RL policy trained without teacher supervision in terms of the reward recieved as well as the average number of goals reached in one rollout. We also observe that our student model is able to reach the performance of the teacher model in fewer iterations and eventually outperforms the teacher.}
    \vspace{-1.5em}
    \label{fig:student-teacher-training}
\end{figure}

\subsection{In-Hand Reorientation Subtask}

In this subtask, the robot hand needs to reorient the tool from an initial orientation to the desired orientation. The tool is initialized with a random orientation in the robot's grasp and the desired orientation is generated at random. When the robot reaches the desired orientation ($\delta \theta < 0.1 rad$), a new goal orientation is generated and the robot needs to move the tool to the next orientation. The performance of the robot is evaluated based on the number of orientations reached in a single episode and the total reward received. We compare the performance of our proposed teacher-student model  (\textbf{S-T}) where the student is pretrained with real world demonstrations to that of the teacher model and a policy that is trained with sparse observation space (same as student) and without teacher supervision (\textbf{RL}). Figure \ref{fig:student-teacher-training} shows the training curves for the in-hand reorientation task. We observe that, during training, our proposed teacher-student framework (\textbf{S-T}) outperformed the teacher policy and the policy trained using \textbf{RL} and sparse rewards in terms of both the reward received as well as the number of goals reached in one episode. During evaluation, we observe that the student model with pretraining and teacher supervision (\textbf{S-T}) outperforms the model trained with sparse data and without teacher supervision (\textbf{RL}) (see Figure \ref{fig:results}). We observe that pretraining from the demonstrations and the supervision from the teacher helps the student to achieve a higher success rate in a shorter training time as compared to the other models. This supports our hypothesis that a teacher-student framework with pretraining will lead to a better performance as compared to \textbf{RL} while achieving shorter training time. 

\subsection{Long-Horizon Task}
\begin{figure}
    \centering
    \includegraphics[width=0.7\columnwidth]{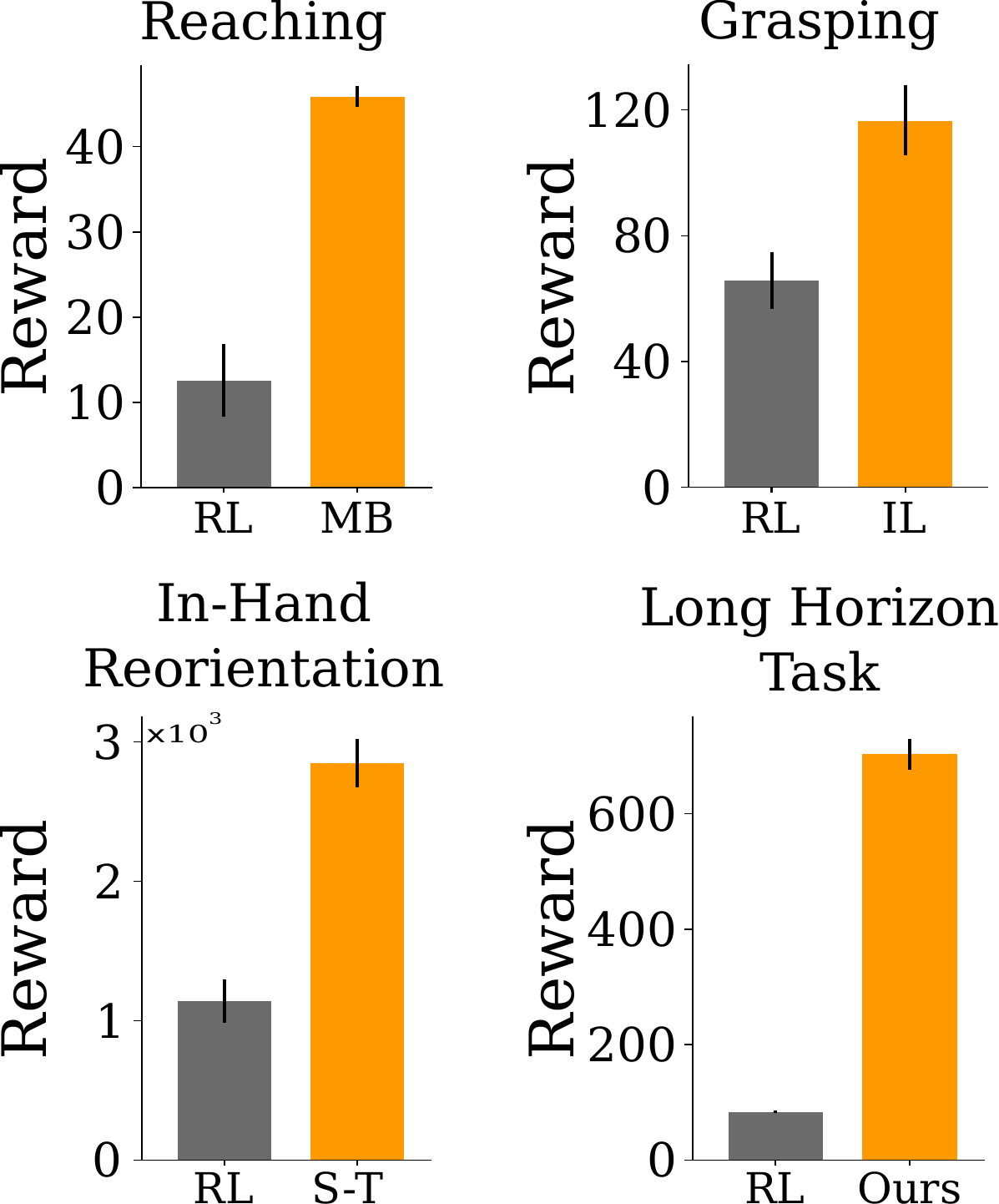}
    \caption{Evaluation results for simulation experiments. We observe that across all subtasks and the long horizon task, using our suggested approach the robot is able to achieve a higher reward as compared to using only RL based approaches. }
    \vspace{-1.5em}
    \label{fig:results}
\end{figure}

Finally, we combine the reaching, grasping, and in-hand reorientation tasks and design a long-horizon task where the robot hand needs to reach for the tool, grasp the tool and position it in a stable grasp, re-orient the tool within $0.1 rad$ of desired orientation and then carry the tool within $0.02m$ of the tool use location. We compare the performance of our unified approach (\textbf{Ours}) where we combine different strategies for solving different subtasks to that of an approach with low-level policies trained using reinforcement learning (\textbf{RL}). We evaluate the performance of this task and report the sum of the reward achieved for each segment and the overall success of the task averaged over $25$ rollouts. We observe that the baseline policy was unable to solve for the complete task and often failed to transition after the first subtask of reaching.
Since the policies trained using reinforcement learning did not perform as well for individual subtasks (see sections A-C), the high level policy could not solve for the entire task. On the other hand, using \textbf{Ours}, the robot was able to complete the task with a success rate of $100\%$. From Figure \ref{fig:results}, we also observe that \textbf{Ours} received a higher reward as compared to the baseline.

\subsection{Real world Deployment}
Now that we have tested the performance of our proposed approach for long horizon task in the simulations, we move on to deploy the unified framework in the real world. We measure the performance of the robot in terms of success rate, where we count the task as successful if the robot grasps and picks up the tool and carries it within $0.02m$ of a randomly sampled desired location within a $0.25\times0.25\times0.5 m$ cube. The robot was initialized at a random position and orientation in its workspace and the tool (weighing $150 g$) was randomly placed in a $10\times 10 cm$ square on the table. The results suggest that on deployment in the real world, our proposed unified framework achieved a success rate of $83.33 \%$ averaged over 12 rollouts. This supports our hypothesis that through a unified policy using different approaches to solve different subtasks of a long horizon task, the autonomous agent learns to perform the task with a high success rate in the simulations and can transfer the policy from the simulation to the real world successfully.
\section{Conclusion}
In this paper, we tackle the problem of solving for long horizon task-oriented dexterous manipulation with anthropomorphic robot hands. We break down the long horizon task into smaller subtasks, and propose a framework that considers available task information, feasibility and level of human effort required for defining the models, providing demonstrations, and designing the reward functions to select the approach best suited for solving each subtask. 
We highlight the use of an Imitation Learning framework that leverages ensemble of policies and look-ahead to learn the dexterous manipulation tasks. We then propose a teacher-student framework that incorporates real world data into the student's training. Finally, we unify all the methods used to solve the different subtasks under a higher level policy. In simulation, we show that our proposed approaches for solving for each subtask outperforms reinforcement learning. Our unified framework leads to a successful transition between different policies and receives a higher reward and success rate. Finally, we show that our proposed approach can be transferred to the real world successfully. This shows the advantage of combining different approaches under a unified policy.
With this work we highlight the feasibility of combining different approaches under a unified framework for long-horizon dexterous manipulation tasks, however this line of work has scope of development in the future works.


\section*{Acknowledgment}
Authors would like to thank Jinda Cui for his immense support with robot teleoperation pipeline and insights on data collection for imitation learning.

\vspace{-0.3em}
\bibliographystyle{IEEEtran}
\bibliography{IEEEexample}

\end{document}